\documentclass[11pt,letterpaper]{article}
\usepackage[margin=1in]{geometry}

\usepackage{amsmath,amsfonts,amssymb,amsthm,mathtools}
\usepackage{algorithm,algorithmic}
\usepackage{enumitem}
\usepackage{xspace,xcolor}
\usepackage{comment}
\usepackage{subcaption}
\usepackage{tikz}
\usetikzlibrary{positioning}
\tikzstyle{agent}=[circle,draw,minimum size=0.6cm,inner sep=0pt]
\tikzstyle{envyEdge}=[-latex,line width=0.03cm]
\tikzstyle{eqEdge}=[-latex,line width=0.03cm]
\tikzset{node distance=2cm,on grid}
\usepackage[normalem]{ulem}
\usepackage[square]{natbib} 

\theoremstyle{definition}

\newif\ifsubmission
\submissionfalse 

\newcommand{\tabincell}[2]{\begin{tabular}{@{}#1@{}}#2\end{tabular}}
\usepackage{enumitem}
\usepackage{array}
\usepackage{multirow}
\usepackage{booktabs}
\title{Emotional Conversation Generation with Heterogeneous Graph Neural Network\thanks{A preliminary version appeared in Proceedings of the 35th AAAI Conference on Artificial Intelligence (AAAI)~\citep{liang2020infusing}.
Compared to the conference version, this journal version adds the fifth knowledge (\emph{i.e.}, the audio) and the relationship among them through the manuscript, and includes automatic evaluation and human evaluation for speakers' personalities that measure whether the generated response expresses the same personal style with the dialogue history or not (subsection 3.4 and section 4). We also conduct such experiments based on our model that inputs golden emotion labels (subsection 5.3) and integrates powerful decoders (subsection 5.4). Furthermore, we implement our approach based on the pre-trained language model BART~\citep{lewis-etal-2020-bart} to demonstrate the effectiveness of the proposed method.
}}

\author{
	Yunlong Liang\textsuperscript{\rm 1}\thanks{Work was done when Liang and Zhang were interning at Pattern Recognition Center, WeChat AI, Tencent Inc, China.},
	Fandong Meng\textsuperscript{\rm 2},
	Ying Zhang\textsuperscript{\rm 1},
	Yufeng Chen\textsuperscript{\rm 1},
	Jinan Xu\textsuperscript{\rm 1}\thanks{Jinan Xu is the corresponding author.} \ and Jie Zhou\textsuperscript{\rm 2}\\
	\textsuperscript{\rm 1}Beijing Key Lab of Traffic Data Analysis and Mining,\\
	Beijing Jiaotong University, Beijing, China \\
	\textsuperscript{\rm 2}Pattern Recognition Center, WeChat AI, Tencent Inc, China\\
	\{yunlongliang, zhying, chenyf, jaxu\}@bjtu.edu.cn,\\ \{fandongmeng, withtomzhou\}@tencent.com\\
}
\date{}

\begin{document}
\maketitle

\begin{abstract}
The successful emotional conversation system depends on sufficient perception and appropriate expression of emotions. In a real-life conversation, humans firstly instinctively perceive emotions from multi-source information, including the emotion flow hidden in dialogue history, facial expressions, audio, and personalities of speakers. Then, they convey suitable emotions according to their personalities, but these multiple types of information are insufficiently exploited in emotional conversation fields. To address this issue, in this paper, we propose a heterogeneous graph-based model for emotional conversation generation. Firstly, we design a \emph{Heterogeneous Graph-Based Encoder} to represent the conversation content (\emph{i.e.}, the dialogue history, its emotion flow, facial expressions, audio, and speakers' personalities) with a heterogeneous graph neural network, and then predict suitable emotions for feedback. Secondly, we employ an \emph{Emotion-Personality-Aware Decoder} to generate a response relevant to the conversation context as well as with appropriate emotions, through taking the encoded graph representations, the predicted emotions by the encoder and the personality of the current speaker as inputs. Experiments on both automatic and human evaluation show that our method can effectively perceive emotions from multi-source knowledge and generate a satisfactory response. Furthermore, based on the up-to-date text generator BART, our model still can achieve consistent improvement, which significantly outperforms some existing state-of-the-art models.

\end{abstract}

\newpage
\section{Introduction}
\begin{figure}[h]
    \centering
    \includegraphics[width=0.8\textwidth]{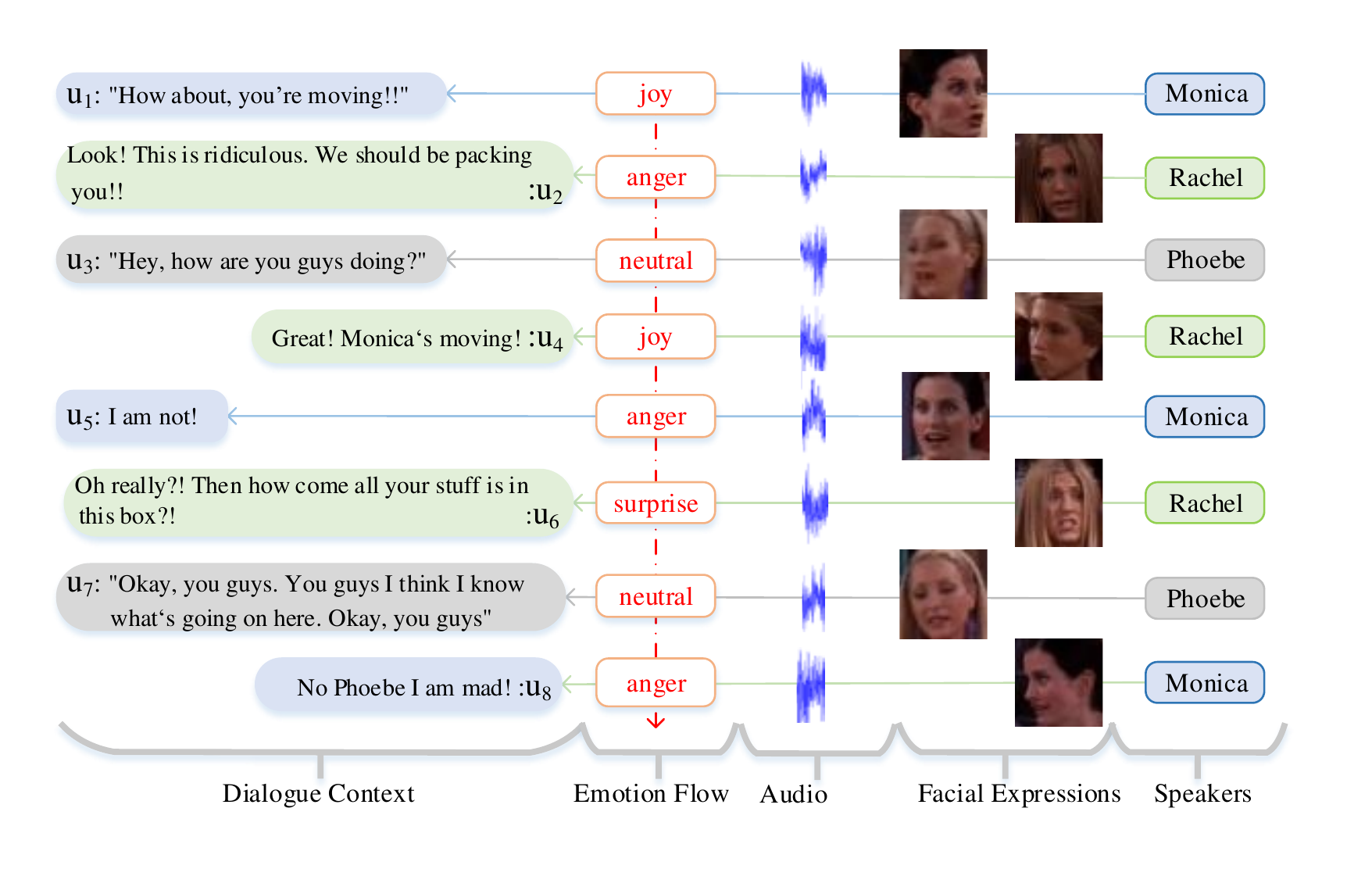}
    \caption{A dialogue example with multi-source knowledge (\emph{i.e.}, the dialogue history, its emotion flow, facial expressions, audio, and speakers). $u_i$: the $i$-th utterance.
    }
    \label{fig:case}
\end{figure}

Injecting emotions into conversation systems can substantially improve its usability and promote customers' satisfaction~\citep{4307122ebb22489cb3a5b11205f1c168,interventionsPartala}. Moreover, perceiving emotions sufficiently is the core premise of expressing emotions~\citep{mayerEI}. In real-world scenarios, humans often instinctively perceive complex or subtle emotions from multiple aspects, including the emotion flow of dialogue history, facial expressions, audio, and personalities of speakers, and then reply with suitable emotions for feedback. Figure~\ref{fig:case} shows the organization of multi-source information in a dialogue graph and the relationship between them. 

In the literature, there have been some efforts in emotional conversation systems, which can be roughly divided into two categories. (1) Several studies focus on building emotion-controllable conversation systems~\citep{zhou2018emotional,colombo-etal-2019-affect,song-etal-2019-generating,shen-feng-2020-cdl} based on user-input emotion, which mainly apply various mechanisms to infuse the specific emotion vector into the response generation process and thus enhance emotional expression. 
However, to some extent, it is a constrained scenario due to requiring an additional emotion input and these methods ignore the information from facial expressions, audio, and speakers' personalities~\citep{madotto-etal-2019-personalizing}. 
(2) Others pay attention to automatically tracking the emotional state in dialogue history to generate emotional responses.\footnote{In our work, we expect to construct a practical and automatic emotional conversation system rather than the emotion-controllable conversation system, so we mainly focus on the second line of work.} For instance,~\citet{Lubis2018ElicitingPE} and~\citet{li2019empgan} utilize an additional RNN-based encoder~\citep{hochreiter1997long} to encode the discrete emotion label sequence (\emph{i.e.}, emotion flow) and maintain it as the emotional context for the decoder, showing that the emotion flow can improve the emotion quality of responses.
However, such specific emotion label sometimes can not fully express the complex and rich emotions of an utterance said by a speaker. 
For example, utterance $u_3$ in Figure~\ref{fig:case} can be labeled as {\em  neutral}, but its corresponding facial expression of the speaker \emph{Phoebe} reflects an additional emotion of {\em joy}. And sometimes we can not correctly perceive the speaker's emotions only through textual expressions, such as utterance $u_5$ in Figure~\ref{fig:case}, whose emotion (\emph{i.e.}, {\em anger}) can be told by the corresponding facial expression and audio of the speaker \emph{Monica}. 
Besides, the emotion is naturally inseparable from speaker's personality~\citep{zhong2020personabased}. For instance, throughout the entire dialogue in Figure~\ref{fig:case}, utterances said by the speaker \emph{Rachel} always have strong emotions due to her personality created in {\em the Friends} series, even though some of them contain no emotional words. 
Therefore, previous models perceive emotions only from the dialogue history and its emotion flow with discrete emotion labels while neglecting complex emotion flows from speakers' facial expressions, audio, and personalities may limit their performances.

In this work, inspired by the capability of heterogeneous graph neural network on capturing information from various types of nodes and relations~\citep{10.1145/3292500.3330961}, we propose a heterogeneous graph-based model to perceive emotions from different types of multi-source knowledge (\emph{i.e.}, the dialogue history, its emotion flow, facial expressions, audio, and speakers' personalities) and then generate a coherent and emotional response. Our model consists of a \emph{Heterogeneous Graph-Based Encoder} and an \emph{Emotion-Personality-Aware Decoder}. Specifically, we build a heterogeneous graph on the conversation content with five-source knowledge, and conduct representation learning over the constructed graph with the encoder to fully understand dialogue content, perceive emotions, and then predict suitable emotions as feedback. The decoder takes the graph-enhanced representation, the predicted emotions, and the current speaker's personality as inputs, and generates a coherent and emotional response.

To verify the proposed method, we conduct experiments on the emotional conversation benchmark dataset MELD~\citep{poria-etal-2019-meld} to evaluate our model with both automatic and human evaluation. Experimental results show that comparing with several competitive baselines (\emph{e.g.}, one previous state-of-the-art model named ReCoSa~\citep{zhang-etal-2019-recosa}), our model can achieve sufficient emotion perception and generate more relevant responses with appropriate emotions, through infusing multi-source knowledge via the heterogeneous graph neural network.
Furthermore, to evaluate the generalizability of our model, we conduct experiments on the DailyDialog~\citep{li-etal-2017-dailydialog} dataset. Although this dataset only contains knowledge from two sources, \emph{i.e.}, the dialogue history and its emotion flow, our model can still generate more satisfactory responses compared with baselines. Furthermore, based on the up-to-date text generator BART~\citep{lewis-etal-2020-bart}, our model still can achieve consistent improvement.

Our contributions can be mainly summarized as follows:
\begin{itemize}
\item We propose a novel heterogeneous graph-based framework for perceiving emotions from different types of multi-source knowledge to generate a coherent and emotional response. To the best of our knowledge, we are the first to introduce heterogeneous graph neural networks for emotional conversation generation.
\item Experimental results on two datasets suggest that our model yields new state-of-the-art performances in emotional conversation generation after incorporating BART into our system, which demonstrate the generalizability of our model, which can be easily adapted to different number of information sources
\item We have uploaded our code at Github\footnote{https://github.com/XL2248/HGNN} for reproducing the results and future work.
\end{itemize}

\begin{figure*}[ht]
\centering
  \includegraphics[width = 0.97 \textwidth]{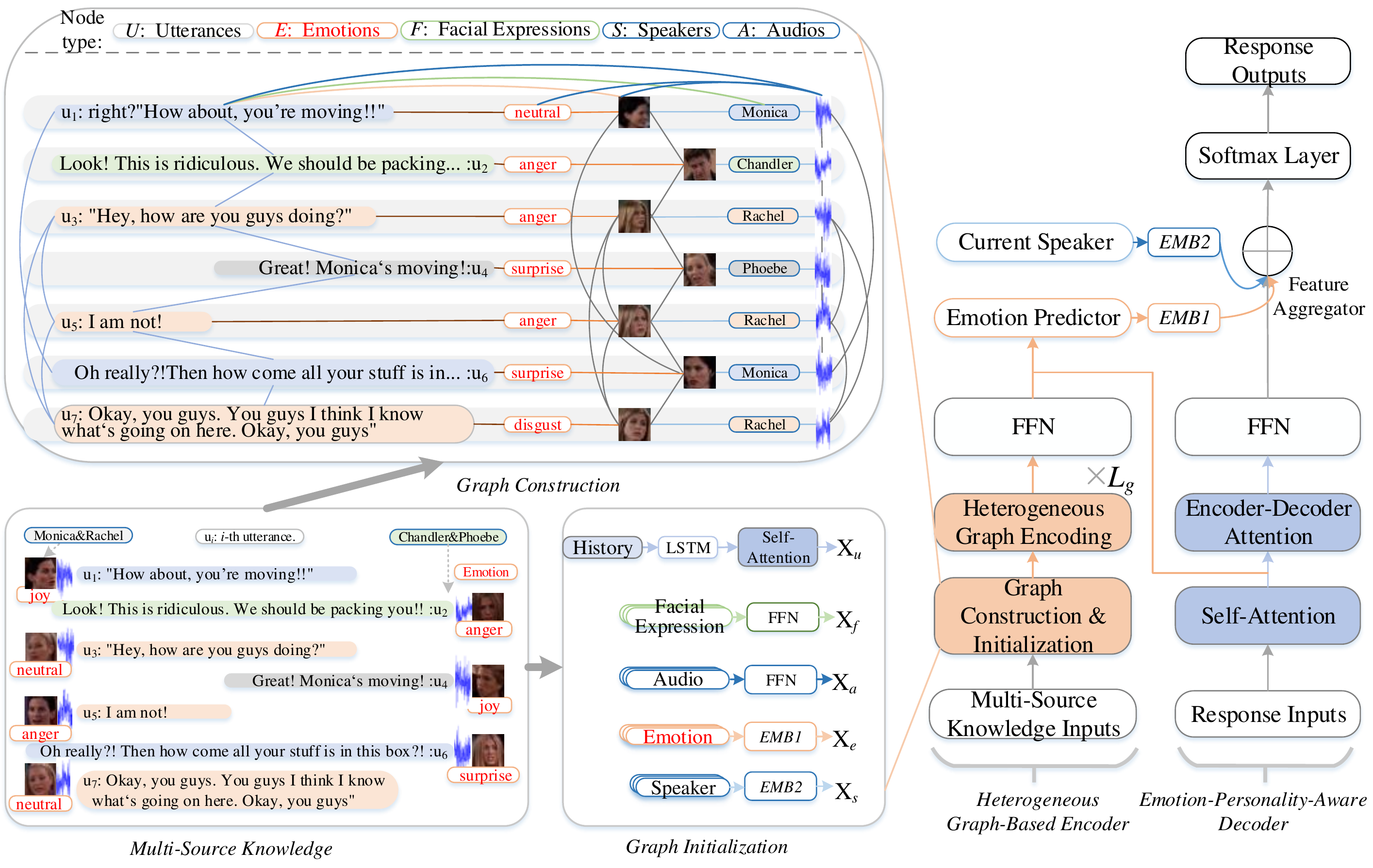}
\caption{The model architecture (on the right) together with an input example (on the left). The model consists of two parts: (1) \emph{Heterogeneous Graph-Based Encoder} to encode multi-source knowledge via heterogeneous graph neural network and (2) \emph{Emotion-Personality-Aware Decoder} to generate a suitable response according to the graph-enhanced representation, the emotions predicted by the encoder and the personality of the current speaker. Note that we only demonstrate the relations among utterance $u_1$, its emotion category, and its associated facial expression/audio/speaker in the `Graph Construction' while omitting those for dialogue rounds $u_2 \sim u_7$ for simplicity.} 
\label{fig:Architecture} 
\end{figure*}

\section{Our Approach}
We elaborate our approach from the following three aspects: Task Definition, Architecture and Training Objective. 

\subsection{Task Definition}
Given the six-tuples $<U$, $F$, $A$, $E$, $S$, $s_{N+1}>$ as inputs, where $U = \{u_1, \cdots, u_N\}$ is the dialogue history till round $N$, $F = \{f_1, \cdots, f_N\}$ is facial expression sequence, $A = \{a_1, \cdots, a_N\}$ is audio sequence, $S = \{s_1, \cdots, s_N\}$ is speaker sequence, $E = \{e_1, \cdots, e_N\}$ is emotion sequence, and $s_{N+1}$ is the speaker for the next response, the goal is to generate a target response $Y = \{y_1, \cdots, y_J\}$ with $J$ words for the ($N$+1)-th round (actually the $u_{N+1}$), which is coherent with the content as well as with appropriate emotions. The dialogue history $U$ contains $N$ utterances, where the $i$-th utterance $u_i = \{x_1, \cdots, x_M\}$ is a sequence with $M$ words.
Corresponding to the $i$-th utterance $u_i$ said by the speaker $s_i$, the facial expression $f_i$ is a vector extracted by the OpenFace~\citep{baltrusaitis2018openface} toolkit\footnote{https://github.com/TadasBaltrusaitis/OpenFace} from the dialogue video\footnote{Each utterance is labeled with a video in MELD dataset. }, the audio $a_i$ is a vector of each utterance trained for emotion/sentiment classification\footnote{These features are originally extracted from openSMILE and then followed by L2-based feature selection using SVM~\citep{poria-etal-2019-meld}.}, and the emotion label $e_i$ is one of seven emotion categories \{\emph {anger, disgust, fear, joy, sadness, surprise, neutral}\}.

\subsection{Architecture}
As shown in Figure~\ref{fig:Architecture}, our architecture contains two components \emph{Heterogeneous Graph-Based Encoder} and \emph{Emotion-Personality-Aware Decoder}, where the encoder aims to understand the content and perceive emotions from conversation context with multi-source information, and the decoder serves for generating a coherent as well as emotional response.
Specifically, our encoder mainly consists of four parts: (1) \emph{Graph Construction}, constructing the heterogeneous graph with multi-source knowledge; (2) \emph{Graph Initialization}, initializing different kinds of nodes;
(3) \emph{Heterogeneous Graph Encoding}, perceiving emotions and representing the conversation context based on the constructed graph; and (4) \emph{Emotion Predictor}, predicting suitable emotions with the graph-enhanced representations for feedback. And our decoder takes the graph-enhanced representation, the predicted emotions, and the current speaker's personality as inputs to generate an appropriate emotional response.

\subsubsection{Heterogeneous Graph-Based Encoder}
\paragraph{Graph Construction.}

As shown in the left part of Figure~\ref{fig:Architecture}, we construct the heterogeneous graph $\mathcal{G} = (\mathcal{V}, \mathcal{E})$, where $\mathcal{V}$ denotes the set of nodes and $\mathcal{E}$ denotes the set of edges represents the relations between nodes. Specifically, we consider five types of heterogeneous nodes, where each node is an utterance in the dialogue history $U$, or a facial expression in $F$, or an audio in $A$, or an emotion category in $E$ or a speaker in $S$. 
Then, we build edges $\mathcal{E}$ among these nodes because there is a natural and close connection between two nodes: 
\begin{enumerate}[itemindent=1em]
\item Between two utterances that are adjacent or said by the same speaker; 
\item Between an utterance and its corresponding facial expression;
\item Between an utterance and its corresponding audio;
\item Between an utterance and its corresponding emotion;
\item Between an utterance and its corresponding speaker who said it;
\item Between two facial expressions that are adjacent or from the same speaker;
\item Between two audios that are adjacent or from the same speaker;
\item Between a facial expression and its corresponding speaker;
\item Between an audio and its corresponding speaker;
\item Between a facial expression and an emotion that correspond to the same utterance;
\item Between an audio and an emotion that correspond to the same utterance.
\end{enumerate}

\paragraph{Graph Initialization.}

We take one dialogue with $N$-round history as an example to describe how to initialize the five types of nodes.

For {\bf utterance nodes}, we firstly use a Long Short-Term Memory (LSTM) network~\citep{Bahdanau2015NeuralMT} to encode each utterance $u_i$ and take the last hidden state of the LSTM as the primary representation $\mathbf{h}_{u_i}$ of the utterance node. By doing so, we obtain the primary representations of $N$ utterance nodes \{$\mathbf{h}_{u_1},\cdots,\mathbf{h}_{u_N}$\}. We next employ multi-head self-attention~\citep{DBLP:journals/corr/VaswaniSPUJGKP17} to generate the contextual representation of each utterance node by aggregating the features from other utterances of the dialogue history. We also use the position embeddings ($\mathbf{PE}$)~\citep{DBLP:journals/corr/VaswaniSPUJGKP17} to distinguish different positions among utterances by concatenating them to the primary utterance representations, \emph{i.e.}, $\mathbf{H}_u$=\{$[\mathbf{h}_{u_1};\mathbf{PE}_N],\cdots,[\mathbf{h}_{u_N};\mathbf{PE}_1]$\}, where we set the position index start from the last history utterance to the first one and $[\cdot;\cdot]$ denotes concatenation operation. 
Finally, we calculate the contextual representations for the $N$ utterance nodes as follows: 
\begin{equation}
  \label{mulitattention}
  \begin{split}
  \mathbf{X}_u &= {\rm MultiHead}(\mathbf{H}_u, \mathbf{H}_u, \mathbf{H}_u), \  \mathbf{X}_u \in \mathbb{R}^{N \times d_u},
  \end{split}
\end{equation}
where ${\rm MultiHead}(\mathbf{Q}, \mathbf{K}, \mathbf{V})$ is a multi-head self-attention function, which takes a query matrix $\mathbf{Q}$, a key matrix $\mathbf{K}$, and a value matrix $\mathbf{V}$ as inputs, and $d_u$ is the dimension of $\mathbf{X}_u$. We take $\mathbf{X}_u$ as the initialized feature of utterance nodes.

For {\bf facial expression nodes}, we firstly extract the original facial expression features\\ $\mathbf{G}_f$=\{$\mathbf{g}_{f_1},\cdots,\mathbf{g}_{f_N}$\}, where $\mathbf{g}_{f_i} \in \mathbb{R}^{d_g}$. Then, we apply a position-wise feed forward network ${\rm FFN}(*)$~\citep{DBLP:journals/corr/VaswaniSPUJGKP17} to project $\mathbf{G}_f$ to textual vector space $\mathbf{X}_f$ as follows:
\begin{equation}
  \label{ffn}
  \begin{split}
  \mathbf{X}_f &= {\rm FFN_f}(\mathbf{G}_f), \  \mathbf{X}_f \in \mathbb{R}^{N \times d_f},
  \end{split}
\end{equation}
where $d_f$ is the dimension of facial expression node representation. We take $\mathbf{X}_f$ as the initialized representation of facial expression nodes.

For {\bf audio nodes}, the original audio feature vectors are handled by~\citep{poria-etal-2019-meld}: $\mathbf{G}_a$=\{$\mathbf{g}_{a_1},\cdots,\mathbf{g}_{a_N}$\}, where $\mathbf{g}_{a_i} \in \mathbb{R}^{d_a}$. Then, we apply a position-wise feed forward network ${\rm FFN}(*)$~\citep{DBLP:journals/corr/VaswaniSPUJGKP17} to project $\mathbf{G}_a$ to textual vector space $\mathbf{X}_a$ as follows:
\begin{equation}
  \label{ffn}
  \begin{split}
  \mathbf{X}_a &= {\rm FFN_a}(\mathbf{G}_a), \  \mathbf{X}_a \in \mathbb{R}^{N \times d_a},
  \end{split}
\end{equation}
where $d_a$ is the dimension of audio node representation. We take $\mathbf{X}_a$ as the initialized representation of audio nodes.

For {\bf emotion nodes} and {\bf speaker nodes}, we maintain trainable parameter matrices $EMB1$ and $EMB2$, which are randomly initialized and can be learned. $EMB1 \in \mathbb{R}^{7 \times d_e}$ are designed for learning seven emotion label features and $EMB2 \in \mathbb{R}^{Z \times d_s}$ are designed for learning the feature of $Z$ speakers' personalities, where $d_e$ and $d_s$ are the dimension of emotion node representation and speaker node representation, respectively. Therefore, the initialized features of emotion nodes and speaker nodes are retrieved from them, denoted as $\mathbf{X}_e$ and $\mathbf{X}_s$, respectively. 

\paragraph{Heterogeneous Graph Encoding.}
Then we introduce the encoding part of the heterogeneous graph neural network (HGNN,~\citep{10.1145/3292500.3330961}) on the constructed heterogeneous graph for capturing features from various types of nodes and relations. 
First, we introduce the conventional adjacency matrix $\mathbf{A} \in \mathbb{R}^{|\mathcal{V}| \times |\mathcal{V}|}$, where $|\mathcal{V}|$ denotes the number of all nodes. The values in $\mathbf{A}$ are either 1 or 0, \emph{e.g.}, $\mathbf{A}_{ij}$=1 denotes there is an edge between node $i$ and node $j$. Then we define five type-wise adjacency matrix $\mathbf{A}_{\tau} \in \mathbb{R}^{|\mathcal{V}| \times |\mathcal{V}|}$, where $\tau \in \{u,f,a,e,s\}$. Specifically, $\mathbf{A}_{\tau}$ is generated by multiplying 
the masked matrix $\mathbf{A}$ with $\mathbf{M}^{\tau} \in  \mathbb{R}^{|\mathcal{V}| \times |\mathcal{V}|}$, where $\mathbf{M}^{\tau}_{i,*}$=1 if the node $i$ is of type $\tau$, otherwise 0. We use $\mathbf{H}^{l} \in \mathbb{R}^{|\mathcal{V}| \times d}$ to denote hidden representations of nodes in the $l$-th layer\footnote{We set $d = d_u = d_f = d_a = d_e = d_s$.}. 

Then, the HGNN considers various types of nodes and maps them into an implicit common space with their individual trainable matrices by:
\begin{equation}
  \label{hgnn}
  \begin{split}
  \mathbf{H}^{l+1} &= \sigma (\sum_{\tau \in \{u,f,a,e,s\}} \mathbf{A}_{\tau}\mathbf{H}^{l}\mathbf{W}^{l}_{\tau}+\mathbf{b}^{l}_{\tau}),
  \end{split}
\end{equation}
where $\sigma$ is an activation function (\emph{e.g.}, ReLU). $\mathbf{H}^{l+1}$ is obtained by aggregating features of their neighboring nodes $\mathbf{H}^{l}$ with different type node $\tau$ using its corresponding transformation matrix $\mathbf{W}^{l}_{\tau} \in \mathbb{R}^{d \times d}$ and bias $\mathbf{b}^l_{\tau} \in \mathbb{R}^{d}$. Initially, $\mathbf{H}^{0} =[\mathbf{X}_{u};\mathbf{X}_{f};\mathbf{X}_{a};\mathbf{X}_{e};\mathbf{X}_{s}]$. By stacking ${L_g}$ such layers, HGNN can aggregate features from various nodes, and the $\mathbf{H}^{L_g}$ contains the representations of all nodes. Then, we get the final output $\mathbf{H}^{enc}$ by:
\begin{equation}
  \label{ffn_o}
  \begin{split}
  \mathbf{H}^{enc} &= {\rm FFN}(\mathbf{H}^{L_g}), \ \mathbf{H}^{enc} \in \mathbb{R}^{|\mathcal{V}| \times d}.
  \end{split}
\end{equation}

If we ignore the heterogeneity of different node types, 
the HGNN will become a conventional homogeneous graph neural network~\citep{kipf2017semi}:
\begin{equation}
  \label{GNN}
  \begin{split}
  \mathbf{H}^{l+1} &= \sigma (\mathbf{A}\mathbf{H}^{l} \mathbf{W}^{l}+\mathbf{b}^{l}),
  \end{split}
\end{equation}
where $\mathbf{W}^{l}$ and $\mathbf{b}^{l}$ are layer-wise trainable weights independent of node types. 

\paragraph{Emotion Predictor.}
\label{sec:emotionclassifier}
After fully perceiving emotions from multiple different sources, the encoder stores the representation in the final layer. We transform the representation $\mathbf{H}^{enc}$ into a fixed-size vector through $\rm{meanpooling}$ operation. Then we apply a fully-connected layer to predict suitable emotions:
\begin{equation}%
  \label{emotion_softmax1}
  \begin{split}
  \mathbf{H}^{mean} &= {\rm Meanpooling}(\mathbf{H}^{enc}),\ \mathbf{H}^{mean} \in \mathbb{R}^{d}, 
  \end{split}
\end{equation}
\begin{equation}%
  \label{emotion_softmax}
  \begin{split}
    \mathcal{P} &= {\rm Softmax}(\mathbf{W}^p\mathbf{H}^{mean}), 
  \end{split}
\end{equation}
where $\mathbf{W}^p \in \mathbb{R}^{7 \times d}$ is trainable parameters\footnote{We omit the bias in {\rm Softmax} layer, similarly hereinafter.}.

\subsubsection{Emotion-Personality-Aware Decoder.}
We aim to incorporate multi-source knowledge with heterogeneous graph neural network for emotional perception, so we use a simple and general decoder, \emph{i.e.} the self-attention based decoder~\citep{DBLP:journals/corr/VaswaniSPUJGKP17}, to generate the response word by word, as shown in the right part of Figure~\ref{fig:Architecture}. Obviously, a more powerful emotion-aware decoder can be extended to our framework, such as~\citep{zhou2018emotional,song-etal-2019-generating}, which is also verified in subsection 4.3.4.

For generating the $t$-th word $y_t$, we firstly take the previous words $y_{1:t-1}$ as inputs to get the representation with a multihead self-attention (future masked) as follows\footnote{We omit the layer normalization for simplicity, and you may refer to~\citet{DBLP:journals/corr/VaswaniSPUJGKP17} for more details.}:
\begin{equation}
  \label{mulitattention_r}
  \begin{split}
  \mathbf{H}^r &= {\rm MultiHead}(\mathbf{R}, \mathbf{R}, \mathbf{R}),
  \end{split}
\end{equation}
where $\mathbf{R}$ is the embedding sequence of the already generated words of target response $r$, \emph{i.e.}, $y_{1:t-1}$.

Then, we use another multi-head attention followed by an ${\rm FFN}$ layer, taking the response history representation $\mathbf{H}^{r}$ as query, and taking $\mathbf{H}^{enc}$ as key and value to output the representation $\mathbf{O} \in \mathbb{R}^{(t-1) \times d}$:
\begin{equation}
  \label{mulitattention-o}
  \begin{split}
  \mathbf{O} &= {\rm FFN}({\rm MultiHead}(\mathbf{H}^r, \mathbf{H}^{enc}, \mathbf{H}^{enc})).
  \end{split}
\end{equation}

To effectively incorporate the predicted emotions and the speaker's personality into the generation process, we design a gate to dynamically control the contribution of the information:
\begin{equation}
  \label{gate1}
  \begin{split}
  \mathbf{O}^{es} &=  \mathbf{O} + g \odot \mathbf{E}_g  + (1-g) \odot \mathbf{S}_g, 
  \end{split}
\end{equation}
\begin{equation}
  \label{gate2}
  \begin{split}
  {g} &= \sigma([\mathbf{O}; \mathbf{E}_g; \mathbf{S}_g]\mathbf{W}^g+\mathbf{b}^g), 
  \end{split}
\end{equation}
\begin{equation}
  \label{gate}
  \begin{split}
  \mathbf{E}_p &= \sum(\mathcal{P} \cdot EMB1), \ \mathbf{E}_p \in \mathbb{R}^{d},
  \end{split}
\end{equation}
where $\mathbf{E}_p$ denotes the mixed emotional representation generated by the weighted sum of the emotion distribution $\mathcal{P}$ (predicted by the encoder as Eq.~\ref{emotion_softmax}) and the emotion parameter matrix $EMB1$. The $\mathbf{S}_p$ is the feature of current speaker $s_{N+1}$ retrieved from the speaker parameter matrix $EMB2$. We repeatly concatenate $\mathbf{E}_p$ for $(t-1)$ times, \emph{i.e.}, $\mathbf{E}_g \in \mathbb{R}^{(t-1) \times d} = [\mathbf{E}_p;\cdots;\mathbf{E}_p]$, similarly, $\mathbf{S}_g \in \mathbb{R}^{(t-1) \times d}= [\mathbf{S}_p;\cdots;\mathbf{S}_p]$. $\mathbf{W}^g \in \mathbb{R}^{d \times d}$ and $\mathbf{b}^g \in \mathbb{R}^{d}$ are the trainable parameters. 

Finally, a softmax layer is utilized to obtain the word probability by taking the emotion-personality-aware representation $\mathbf{O}^{es}$ as input. Therefore, the probability of word $y_t$ is calculated as follows:
\begin{equation}
  \label{log_}
  \begin{split}
  P(y_t|y_{1:t-1};\mathcal{G};\mathbf{E}_p;\mathbf{S}_p;\theta) &= {\rm Softmax}(\mathbf{W}^o\mathbf{O}^{es}_t),
  \end{split}
\end{equation}
where $\mathcal{G}$ is the constructed graph; $\mathbf{W}^o \in  \mathbb{R}^{|V| \times d}$   is trainable parameter, where $V$ is the vocabulary size. The log-likelihood of the corresponding response sequence ${Y} = \{ y_1, \cdots, y_J\}$ is:
\begin{equation}
\begin{split}
  P({Y}|\mathcal{G};\mathbf{E}_p;\mathbf{S}_p;\theta) &= \prod_{t}P(y_t|y_{1:t-1};\mathcal{G};\mathbf{E}_p;\mathbf{S}_p; \theta).
\end{split}
\end{equation}

\subsection{Training}
Our model can be trained end to end, and the overall training objective consists of the response generation loss $\mathcal L_{MLL}$ and the emotion classification loss $\mathcal L_{CLS}$ as follows:
\begin{equation}
\begin{split}
  \mathcal{J} &= (1 - \lambda)\mathcal L_{MLL} + \lambda\mathcal L_{CLS},
  \label{final_loss}
\end{split}
\end{equation}
\begin{equation}
\begin{split}
  \mathcal L_{MLL} &=  \min (- \sum_{t=0}^{J} \log({y}_{t})), 
\end{split}
\end{equation}
\begin{equation}
\begin{split}
  \mathcal L_{CLS} &=  \min (- \log(\mathcal{P}[e_{N+1}])),
\end{split}
\end{equation}
where $\lambda$ is the discount coefficient to balance the generation loss and classification loss. And $e_{N+1}$ is the golden class label, $\mathcal{P}$ is calculated as in Eq.~\ref{emotion_softmax}.

\section{Experiments}
\label{sec:length}
\subsection{Datasets}
We conduct experiments on the following two datasets to verify the performance of our model: MELD and DailyDialog.

\subsubsection{MELD} MELD~\citep{poria-etal-2019-meld} is a multi-party and multi-modal emotional dialogue dataset from the Friends TV series, which contains textual, acoustic, video, and speakers information. We preprocess each video into a sequence of images with the speaker’s facial expression according to the video resolution. Then we use the OpenFace toolkit~\citep{baltrusaitis2018openface} to extract the facial features from each image and average these features across the images of the video as the final expression for the corresponding utterance, where we only use the feature of facial action units for emotional response generation. And the audio has been preprocessed to the feature vector with 1611/1422 dimensional size of each utterance for emotion/sentiment classification by~\citep{poria-etal-2019-meld}, where we used the 1611 dimensional size for emotional conversation generation. Each utterance in every dialogue is labeled with one of the seven emotion categories. Moreover, the MELD can be used for emotion recognition~\citep{majumder2019dialoguernn,Qin_Che_Li_Ni_Liu_2020,qin2021co} and recently is also extended to conduct chat translation~\citep{liang-etal-2021-modeling,liang-etal-2021-towards}, showing the wide applicability of the dataset.

\subsubsection{DailyDialog} 
To verify the generalizability of our model across datasets, we conduct experiments on DailyDialog~\citep{li-etal-2017-dailydialog}, a larger scale dataset only containing textual utterances with the same emotion categories as MELD. Therefore, we remove the facial expression nodes and the speaker nodes from the constructed graph $\mathcal{G}$ to adapt our model to DailyDialog. The statistics of emotion distribution and training, validation, and test splits are shown in Table~\ref{datasets}. 

\begin{table}[h]
\centering

\begin{tabular}{l|rrr|rrr}
\toprule
\multirow{2}{*}{Categories} & \multicolumn{3}{c|}{MELD} &  \multicolumn{3}{c}{DailyDialog} \\\cline{2-7}
&Train &Validation & Test&Train &Validation & Test\\\hline
\# anger  &1,109&153&345  &827&77&118 \\
\# disgust  &271&22&68  &303&3&47 \\
\# fear &268&40&50   &146&11&17 \\
\# joy &1,743&163&402   &11,182&684&1,019 \\
\# neutral &4,710&470&1,256   &72,143&7,108&6,321 \\
\# sadness &683&111&208&969&79&102 \\
\# surprise &1,205&150&281&1,600&107&116 \\\hline
\# dialogues &1,039&114&280&11,118&1,000&1,000 \\
\# utterances &9,989&1,109&2,610&87,170&8,069&7,740 \\
\bottomrule
\end{tabular}
\caption{Statistics of MELD and DailyDialog.
}\label{datasets}
\end{table}

\subsection{Comparison Models}
We compare our model with the following baselines:
\begin{itemize}
\item \textbf{Seq2Seq}~\citep{sutskever2014sequence}: It is the simplest generation-based approach, which has been widely applied to generation tasks.

\item \textbf{HRED}~\citep{DBLP:journals/corr/SerbanSBCP15}: It is a hierarchical encoder-decoder network, which performs well due to its context-aware modeling ability. 

\item \textbf{Emo-HRED}~\citep{Lubis2018ElicitingPE}: It is an extension of HRED by adding an additional encoder to represent the emotional labels of dialogue as the emotional context for the decoder. 

\item \textbf{ReCoSa}~\citep{zhang-etal-2019-recosa}\footnote{https://github.com/zhanghainan/ReCoSa}: It uses self-attention to measure the relevance between response and dialogue history, and shows state-of-the-art performances on benchmark datasets. In order to make the comparison more convincing, we extend this model to utilize multi-source knowledge by concatenating other source features (\emph{i.e.}, facial expressions, emotions, and speakers' personalities) to dialogue history representations, and train it with the same loss function $\mathcal{J}$. Besides, we replace its original decoder with \emph{Emotion-Personality-Aware Decoder} for fair comparison.

\item \textbf{BART}~\citep{lewis-etal-2020-bart}\footnote{We use the bart large setting: {https://github.com/pytorch/fairseq/tree/main/examples/bart}}: It is the current up-to-date text generators pre-trained on large-scale text data. We implement our approach based on it to further demonstrate the effectiveness of our approach under the pre-trained language model setting. For a fair comparison, we also incorporate the additional source knowledge into BART by concatenating these knowledge under different settings.

\item \textbf{GNN}~\citep{kipf2017semi}: We adapt our architecture to a conventional graph-based network by taking all nodes as homogeneous ones.

\end{itemize}

\subsection{Settings and Hyperparameters}

For a fair comparison, we train our models with the same settings as comparison models. Specifically, the word embedding dimension and hidden size are set to 128 and 256, respectively. The $d_u, d_f, d_e, d_s$ and $d$ are set to 256. The number of encoder layers $L_g$ is 2. The number of attention heads of ReCoSa and our model are set to 4. The discount coefficient $\lambda$ is empirically set to 0.5. The dropout rate is set to 0.1 and the batch size is 16. Adam~\citep{Adam:14} is utilized for optimization with the learning rate 0.001. The trainable parameter matrices $EMB1$ and $EMB2$ are randomly initialized by Xavier~\citep{pmlr-v9-glorot10a}. The max turn of dialogue is 35 and the max sentence length is 50. We set the number of speakers $Z$ to 13. Since most speakers just said several utterances in the whole corpus, to learn better representations of speakers' personality, we remove the speakers whose utterances are less than 30 pieces and finally retain 12 speakers. To adapt a new speaker, we add $UNK$ for unknown speakers and there are total 13 speakers. For the MELD and DailyDialog, we follow ReCoSa~\citep{zhang-etal-2019-recosa} and set the vocabulary size (\emph{i.e.}, $V$) to 5,253 and 17,657, respectively. When implementing our model based on BART~\citep{lewis-etal-2020-bart}, we added the heterogeneous graph neural network and an emotion predictor on the BART encoder, where the $d_u, d_f, d_e, d_s$ and $d$ are set to 768. When training, the parameters of the heterogeneous graph neural network and emotion predictor are randomly initialized and the rest of the parameters are initialized with the pre-trained BART. Then, all parameters are updated with the training objective in Eq.~\ref{final_loss}.

\subsection{Evaluation Metrics}
\label{sec:evaluation_metrics}
We adopt the following two metrics to evaluate the model performance, \emph{i.e.}, automatic metrics and human metrics.
\subsubsection{Automatic Metrics}
For measuring content quality, we use PPL and BLEU scores~\citep{xing2017topic} to evaluate the fluency and relevance of generated responses, respectively. We use Dist-1 and Dist-2~\citep{li-etal-2016-deep} to evaluate the degree of diversity. We use three embedding-based metrics~\citep{liu-etal-2016-evaluate}(average, greedy and extreme) to evaluate the semantic-level similarity between the generated responses and the ground truth. They are all widely applied in response generation~\citep{TianContext,10.1145/3178876.3186077,xing2018hierarchical}.

To measure the accuracy of emotion expression for generated responses, we follow~\citep{zhou2018emotional,song-etal-2019-generating} and fine-tune a BERT-Based~\citep{bert} emotion classifier on the text utterances of MELD  
as one evaluation tool. The `W-avg.' result of the evaluation tool is 62.55\%, which is better than existing best classifier DialogueRGAN (60.91\%, also trained on MELD by~\citet{ishiwatari-etal-2020-relation}). The weighted-average F1-score~\citep{ghosal-etal-2019-dialoguegcn} (denoted as W-avg. for short) is the agreement between the predicted label by the evaluation tool and the ground-truth. 

For evaluating personality quality, we follow~\citep{wu-etal-2020-guiding} and adopt uPPL, uDist-1, and uDist-2 to measure the language style of speaker, the diversity between speakers, respectively. To conduct the uPPL metric, given a speaker $s_i$, a statistical language model $LM_i$ is first trained using the speaker’s utterances, which quantifies the power of a persona-aware model on generating responses similar to speakers’ history utterances. Then its corresponding uPPL is defined as the
perplexity of a generated response $r'$ given by $LMi$. For uDist-1 and uDist-2, given a dialogue history $c_i$ and
$m$ different speakers \{$s_j | j \in [1, m]$\}, we generate
different responses \{$r'_j |j \in [1, m]$\} for each speaker
using our model. On this basis, uDist-1 and uDist-2 of the response set \{$r'_j |j \in [1, m]$\} are utilized to measure the in-group diversity of responses generated by our model within speakers. 

\subsubsection{Human Evaluation}
Following~\citep{zhang-etal-2019-recosa,zhong2019affect}, we sample 100 dialogues from the test set and exploit three annotators to evaluate the content, personality and emotion quality. We then randomly disorder the responses generated by each comparison model. For each sample, annotators are asked to score a response from content level, personality level and emotion level independently: 
\begin{itemize}
\item +2: (content) The response is grammatically correct and relevant / (emotion) The response conveys accurate and appropriate emotion / (personality) The response expresses the personality style consistent with the current speaker and suitable to the dialogue history.
\item +1: (content) The response is grammatically correct but irrelevant / (emotion) The response conveys inaccurate but appropriate emotion for the dialogue history / (personality) The response expresses the personality style consistent with the current speaker but not suitable to the dialogue history.
\item 0: (content) The response has incorrect grammar and is irrelevant / (emotion) The response conveys inaccurate and inappropriate emotion / (personality) The response expresses the personality style inconsistent with the current speaker and not suitable to the dialogue history.
\end{itemize}

\section{Results}
We mainly present the models' results from the following two aspects: automatic and human evaluation.
\begin{table}[ht!]
\centering
\setlength{\tabcolsep}{0.20mm}{
\begin{tabular}{l|l|rccccc|c|cc}
\toprule
&\multirow{2}{*}{Models} & \multicolumn{6}{c}{Content}& \multicolumn{1}{c}{Emotion}& \multicolumn{2}{c}{Personality}  \\\cline{3-11}
& &PPL$\downarrow$ & BLEU$\uparrow$ & Dist-1/2$\uparrow$  &Avg.$\uparrow$ &Ext.$\uparrow$ &Gre.$\uparrow$ & W-avg.$\uparrow$ & uPPL$\downarrow$ & uDist-1/2$\uparrow$ \\\hline
\multirow{4}{*}{\tabincell{c}{One.}}
&Seq2Seq &125.29 &1.14 &.063/1.21 &.532&.361&.373&27.88     &188.32 &0.067/1.34 \\
&HRED &121.51 &1.77 &.086/1.55  &.566&.362&.410&28.56     &181.51 &0.097/1.75  \\
&ReCoSa  &114.19 &2.11 &1.38/2.57 &.565&.370&.413 &28.96    &169.34 &0.95/2.27 \\ 
&BART  &21.36 &3.19 &3.40/6.32 &.588&.392&.432 &33.93    &74.45 &3.88/7.52 \\ \hline
\multirow{6}{*}{\tabincell{c}{Two.}}
&Emo-HRED  &116.14 &1.87 &1.23/3.77 &.566&.369&.412&29.67   &171.14 &1.15/3.12 \\
&ReCoSa &108.51 &2.00 &1.22/4.54 &.568&.365&.419&30.13      &160.92 &1.27/3.43 \\
&BART &21.24 &3.14 &3.42/7.41 &.593&.401&.441 &33.17    &74.11 &5.15/8.01 \\
&GNN (Ours)  &108.47 &2.06 &1.75/2.58 &.571&.364&.419 &30.18    &{159.78} &{1.66}/{2.52}  \\
&HGNN (Ours)&{106.04} &{1.91}  &2.52/5.95 &.583&.376&.427&31.58   &{156.75} &{1.75}/{3.94} \\
&HGNN+BART (Ours)&{19.98} &3.25 &4.13/7.76 &.609&.409&.453 &34.12    &73.62 &5.59/8.59  \\\hline

\multirow{5}{*}{\tabincell{c}{Four.}}
&ReCoSa &102.37 &2.49 &2.07/6.88  &.571&.364&.421&30.66  &152.88&1.88/5.07 \\
&BART &20.95 &3.37 &5.29/9.41 &.609&.413&.465 &34.58    &73.75 &5.83/8.15 \\
&GNN (Ours) &100.66 &2.15 &2.92/4.63 &.578&.372&.426 &31.25  &151.29&1.90/4.88\\
&HGNN (Ours) &{96.90} &{2.80} &{4.84}/{9.15}  &{.593}&{.394}&{.437}&{34.49}&149.33&2.09/7.43\\
&HGNN+BART (Ours) &{19.09} &3.66 &6.17/9.81 &.618&.422&.474 &34.86    &72.79 &7.05/8.84 \\\hline
\multirow{5}{*}{\tabincell{c}{Five.}}
&ReCoSa &101.78 &2.67 &2.72/7.64  &.575&.367&.426&30.97 &150.44&2.35/6.32\\
&BART &19.62 &3.59 &5.89/9.91 &.615&.428&.477 &35.01    &72.82 &6.99/8.39 \\
&GNN (Ours) &100.01 &2.59 &2.92 /5.71 &.579&.376&.429 &31.59  &150.71&2.77/5.77\\
&HGNN (Ours) &{96.34} &{3.08} &{5.08}/{9.58}  &{.597}&{.398}&{.442}&{34.72}&{148.08}&{6.18}/{8.19}\\
&HGNN+BART (Ours) &\bf{18.44} &\bf{3.89} &\bf{6.11}/\bf{11.12}  &\bf{.631}&\bf{.448}&\bf{.495}&\bf{35.83}&\bf{71.53}&\bf{7.13}/\bf{9.21}\\
\bottomrule
\end{tabular}}
\caption{Automatic evaluation results of generated responses on test sets of MELD. `One.': One Source, `Two.': Two Sources, `Four.': Four Sources, and `Five.': Five Sources. Four settings mean different knowledge is used, namely `One Source' for only dialogue history, `Two Sources' for dialogue history and its emotion flow, `Four Sources' for dialogue history, its emotion flow, facial expressions, and speakers' personalities, and `Five Sources' for all available information. Note: `Avg.', `Ext.', and `Gre.' denote three embedding-based metrics (average, greedy, and extreme, respectively). `W-avg.' denotes weighted-average F1-score of seven emotion categories, which is the agreement between the predicted label by the BERT-Based evaluation tool and the ground-truth. `uPPL', `uDist-1' and `uDist-1' aim to measure the personality of generated responses. Note that we also conduct such experiments where we replace the facial expressions with the audio features in `Four Sources' setting, and we obtain the similar result. Therefore, we only list one scenario that contains the facial expression rather than the audio, similarly hereinafter.
}\label{response_measure} 
\end{table}

\subsection{Automatic Evaluation Results}
From the results in Table~\ref{response_measure}, we see that incorporating additional knowledge indeed helps models to generate a response not only more relevant to the content but also with appropriate emotions, suggesting the necessity of infusing multi-source knowledge, especially by our HGNN. We can conclude that from Table~\ref{response_measure} and Table~\ref{response_measure_d}:

(1) In all four settings (`One Source', `Two Sources', `Four Sources', and `Five Sources'), Table~\ref{response_measure} and Table~\ref{response_measure_d} show that our HGNN consistently surpasses other competitors on most evaluation metrics. This suggests that infusing multi-source knowledge with heterogeneous graph has a positive influence on response generation, and demonstrates the superiority of our model on understanding content and emotion, which yields new state-of-the-art performances.



2) In `Two Sources' setting, our HGNN outperforms the best ReCoSa~\citep{zhang-etal-2019-recosa} on MELD by 2.47$\downarrow$, 1.30$\uparrow$, 1.41$\uparrow$, 0.02$\uparrow$, 0.01$\uparrow$, 0.01$\uparrow$, 1.45$\uparrow$, 4.17$\downarrow$, 0.48$\uparrow$ and 0.51$\uparrow$ on PPL, Dist-1, Dist-2, Avg., Ext., Gre., W-avg., uPPL, uDist-1, and uDist-2, respectively, showing that our HGNN indeed generates more fluent, diverse, coherent, emotional, and personalized responses than baselines. Although our HGNN achieves comparable BLEU scores in contrast to ReCoSa (\emph{i.e.} 1.91 vs. 2.00 on MELD, 0.75 vs. 0.72 on DailyDialog), the human evaluation will further verify the effectiveness of our model on generating relevant responses in next section. Furthermore, we validate our HGNN on a larger dataset DailyDialog and the results show the generalizability of our model.


\begin{table*}[t]
\centering
\setlength{\tabcolsep}{0.90mm}{
\begin{tabular}{l|l|rcclcr|c}
\toprule
\multirow{2}{*}{Data Sources}&\multirow{2}{*}{Models} &  \multicolumn{6}{c}{Content}&\multicolumn{1}{c}{Emotion} \\\cline{3-9}
& & PPL$\downarrow$ & BLEU$\uparrow$ & Dist-1/2$\uparrow$ &Avg.$\uparrow$ &Ext.$\uparrow$ &Gre.$\uparrow$& W-avg. $\uparrow$\\\hline
\multirow{4}{*}{\tabincell{c}{One Source}}
&Seq2Seq &85.29 &0.20 &0.013/0.085  &0.579&0.344&0.394&65.40\\
&HRED &82.02 &0.27 &0.015/0.20  &0.577&0.322&0.399&69.49 \\
&ReCoSa &80.43 &0.74 &0.17/0.23  &0.584&0.353&0.390&68.76\\ 
&BART &9.78 &0.77 &0.23/0.37  &0.624&0.397&0.495&71.27\\ \hline
\multirow{6}{*}{\tabincell{c}{Two Sources}}
&Emo-HRED &81.48 &0.54 &0.12/0.23  &0.598&0.350&0.406&70.49\\
&ReCoSa  &80.28 &0.72 &0.13/0.24  &0.602&0.349&0.435&69.63\\
&BART  &9.21 &0.76 &0.25/0.39  &0.635&0.414&0.516&71.75\\
&GNN (Ours)  &{78.81} &{0.65} &{0.24}/{0.40} &0.617&0.372&0.431&69.93  \\
&HGNN (Ours) &{76.59} &{0.75} &{0.22}/{0.47}  &{0.641}&{0.418}&{0.510}&{71.16} \\
&HGNN+BART (Ours) &\bf{8.81} &\bf{0.83} &\bf{0.28}/\bf{0.58}  &\bf{0.655}&\bf{0.457}&\bf{0.558}&\bf{72.69} \\
\bottomrule
\end{tabular}}
\caption{Automatic evaluation results of generated responses on test set of DailyDialog. Due to no speaker information in this dataset, we do not conduct such experiments to measure the personality quality (\emph{i.e.}, uPPL, uDist-1, and uDist-2).
}\label{response_measure_d} 
\end{table*}

\begin{table}[t]
\centering
\setlength{\tabcolsep}{0.90mm}{
\begin{tabular}{l|l|ccc|c|ccc|c}
\toprule
\multirow{2}{*}{Data Sources}&\multirow{2}{*}{Models} & \multicolumn{4}{c|}{MELD} &  \multicolumn{4}{c}{DailyDialog} \\\cline{3-10}
& &+2 & +1 &0 &Kappa &+2 & +1 &0 &Kappa\\\hline
 \multirow{4}{*}{\tabincell{c}{One Source}}
& Seq2Seq &27.00&43.33&39.67&0.531  &25.33&51.00&23.67 &0.562\\
& HRED  &27.33&31.67&41.00&0.547  &25.67&51.66&22.67&0.504 \\
& BART  &29.67&37.00&33.33&0.521  &29.66&53.67&17.67&0.531 \\
& ReCoSa  &27.33&35.34&37.33&0.472   &27.00&51.00&22.00&0.564 \\\hline 
\multirow{6}{*}{\tabincell{c}{Two Sources}}
& Emo-HRED &27.67&34.00&38.33&0.420   &27.33&{52.34}&20.33&0.478 \\
& ReCoSa &28.67&41.00&30.33&0.459   &29.00&51.00&{20.00}&0.495 \\ 
& BART &33.33&44.67&22.00&0.477   &33.34&53.33&{13.33}&0.467 \\ 
& GNN (Ours) &29.00&43.67&27.33&0.499 &32.33&45.00&22.67&0.474 \\
& HGNN (Ours) &29.67&44.33&26.00&0.467 &{34.33}&{43.67}&{22.00}&0.514 \\
& HGNN+BART (Ours) &35.00&45.00&20.00&0.484 &\bf{36.67}&{47.00}&{16.33}&0.509 \\\hline
 \multirow{5}{*}{\tabincell{c}{Four Sources}}
& ReCoSa &34.66&39.67&25.67&0.482&-&-&- \\
& BART &36.00&43.33&20.67&0.456&-&-&- \\
& GNN (Ours) &35.33&{44.34}&{20.33}&0.518&-&-&- \\
& HGNN (Ours) &{36.00}&{42.00}&{22.00}&0.505&-&-&- \\
& HGNN+BART (Ours) &{38.33}&{45.00}&{16.67}&0.546&-&-&- \\\hline
 \multirow{5}{*}{\tabincell{c}{Five Sources}}
& ReCoSa &37.66&38.67&23.67&0.513&-&-&- \\
& BART &39.00&40.67&20.33&0.505&-&-&- \\
& GNN (Ours) &36.00&{45.00}&{18.00}&0.497&-&-&- \\
& HGNN (Ours) &{38.33}&{40.34}&{21.33}&0.521&-&-&- \\
& HGNN+BART (Ours) &\bf{41.67}&{44.00}&{14.33}&0.519&-&-&- \\
\bottomrule
\end{tabular}}
\caption{Human evaluation on content quality.
}\label{human_evaluation_content} 
\end{table}

(3) Compared with GNN in the same settings, our HGNN always surpasses GNN by large margins on both datasets, especially in `Five Sources' setting (\emph{i.e.}, 3.67$\downarrow$ on PPL, 0.49$\uparrow$ on BLEU, 2.16$\uparrow$ on Dist-1, 3.87$\uparrow$ on Dist-2, 0.018$\uparrow$ on Avg., 0.022$\uparrow$ on Ext., 0.013$\uparrow$ on Gre., 3.13$\uparrow$ on W-avg., 2.63$\downarrow$ on uPPL, 1.55$\uparrow$ on uDist-1, and 2.42$\uparrow$ on uDist-2), demonstrating the effectiveness and superiority of the heterogeneous graph on dealing with the various types of multi-source knowledge. In the `Four Sources' setting, there is a similar trend to `Five Sources' setting.

(4) Especially, for the emotion and personality quality (the last three metrics in Table~\ref{response_measure}), we find that our HGNN model consistently outperforms all baselines in four settings, \emph{e.g.}, the best previous model ReCoSa, in `Five Sources' setting, by 2.36$\downarrow$ on uPPL, 1.97$\uparrow$ on uDist-1, and 1.87$\uparrow$ on uDist-2, respectively. This suggests the ability of our model on generating emotion and preserving the personality style consistent with the current speaker.

(5) Based on the up-to-date text generator BART~\citep{lewis-etal-2020-bart}, our model still achieves consistent improvement, suggesting the effectiveness of model and showing the importance of multi-source knowledge.

\begin{table}[t]
\centering
\setlength{\tabcolsep}{0.9mm}{
\begin{tabular}{l|l|ccc|c|ccc|c}
\toprule
\multirow{2}{*}{Data Sources}&\multirow{2}{*}{Models} & \multicolumn{4}{c|}{MELD} &  \multicolumn{4}{c}{DailyDialog} \\\cline{3-10}
& &+2 & +1 &0 &Kappa &+2 & +1 &0 &Kappa \\\hline
 \multirow{4}{*}{\tabincell{c}{One Source}}
& Seq2Seq &13.00&41.33&45.67&0.637  &10.67&47.33&42.00 &0.608\\
& HRED  &13.67&45.67&40.66&0.548  &10.67&48.00&41.33 &0.494\\
& ReCoSa  &14.33&45.00&40.67&0.638   &12.00&47.33&40.67&0.577 \\
& BART  &17.00&49.00&34.00&0.612   &15.33&51.67&33.00&0.515 \\\hline 
\multirow{6}{*}{\tabincell{c}{Two Sources}}
& Emo-HRED &15.33&40.34&44.33&0.649   &12.33&{48.67}&39.00&0.593 \\
& ReCoSa &15.33&48.67&36.00&0.687   &15.00&41.33&43.67&0.602 \\ 
& BART &19.67&54.00&26.33&0.651   &19.67&47.00&33.33&0.579 \\ 
& GNN (Ours) &15.33&44.34&40.33&0.568 &15.33&46.34&38.33&0.567 \\
& HGNN (Ours) &17.67&41.33&41.00&0.593 &{18.00}&{45.00}&{37.00}&0.627 \\
& HGNN+BART (Ours) &21.00&47.00&32.00&0.557 &\bf{21.33}&{48.67}&{30.00}&0.602 \\\hline
 \multirow{5}{*}{\tabincell{c}{Four Sources}}
& ReCoSa &19.67&{50.66}&29.67&0.604&-&-&-&- \\
& BART &25.00&{54.33}&20.67&0.592&-&-&-&- \\
& GNN (Ours) &22.67&49.00&{28.33}&0.643&-&-&-&- \\
& HGNN (Ours) &{24.67}&{41.00}&{34.33}&0.618&-&-&-&- \\
& HGNN+BART (Ours) &{27.33}&{46.67}&{26.00}&0.602&-&-&-&- \\\hline
 \multirow{5}{*}{\tabincell{c}{Five Sources}}
& ReCoSa &21.67&{50.00}&28.33&0.617&-&-&-&- \\
& BART &29.00&{54.00}&17.00&0.583&-&-&-&- \\
& GNN (Ours) &24.67&47.66&{27.67}&0.627&-&-&-&- \\
& HGNN (Ours) &{27.00}&{40.67}&{32.33}&0.595&-&-&-&- \\
& HGNN+BART (Ours) &\bf{31.33}&{48.67}&{20.00}&0.612&-&-&-&- \\
\bottomrule
\end{tabular}}
\caption{Human evaluation on emotion quality.
}\label{human_evaluation_emotion} 
\end{table}

\begin{table}[t]
\centering
\setlength{\tabcolsep}{0.90mm}{
\begin{tabular}{l|l|ccc|c}
\toprule
\multirow{2}{*}{Data Sources}&\multirow{2}{*}{Methods} & \multicolumn{4}{c}{MELD}  \\\cline{3-6}
& &+2 & +1 &0 &Kappa \\\hline
 \multirow{4}{*}{\tabincell{c}{One Source}}
& Seq2Seq &13.00&40.33&46.67&0.467  \\
& HRED  &10.33&46.67&43.00&0.458  \\
& ReCoSa  &11.33&44.00&44.67&0.468  \\
& BART  &16.00&49.00&35.00&0.471  \\\hline 
\multirow{6}{*}{\tabincell{c}{Two Sources}}
& Emo-HRED &11.33&40.34&48.33&0.469   \\
& ReCoSa &12.00&48.33&39.67&0.487  \\ 
& BART  &18.67&51.33&30.00&0.483  \\
& GNN (Ours) &13.33&45.34&41.33&0.485 \\
& HGNN (Ours) &15.67&43.33&41.00&0.503  \\
& HGNN+BART (Ours) &20.00&50.00&30.00&0.467  \\\hline
 \multirow{5}{*}{\tabincell{c}{Four Sources}}
& ReCoSa &16.67&{50.66}&33.67&0.510\\
& BART  &22.00&50.33&27.67&0.491  \\
& GNN (Ours) &18.00&51.33&{30.67}&0.513\\
& HGNN (Ours) &{20.67}&{45.00}&{34.33}&0.488\\
& HGNN+BART (Ours) &{24.33}&{51.67}&{24.00}&0.521\\\hline
 \multirow{5}{*}{\tabincell{c}{Five Sources}}
& ReCoSa &18.67&{50.33}&31.00&0.512 \\
& BART  &23.67&52.33&24.00&0.497  \\
& GNN (Ours) &20.00&48.33&{30.67}&0.527\\
& HGNN (Ours) &{22.33}&{44.67}&{33.00}&0.502\\
& HGNN+BART (Ours) &\bf{25.00}&{51.00}&{24.00}&0.528\\
\bottomrule
\end{tabular}}
\caption{Human evaluation on personality quality.
}\label{human_evaluation_per} 
\end{table}

\begin{table}[t]
\centering
\setlength{\tabcolsep}{0.90mm}{
\begin{tabular}{c|cccc|cccc}
\toprule
\multirow{2}{*}{Models} & \multicolumn{4}{c|}{MELD} &  \multicolumn{4}{c}{DailyDialog} \\\cline{2-9}
 &Emo-HRED &ReCoSa &GNN &HGNN &Emo-HRED &ReCoSa &GNN &HGNN \\\hline
 Emo-HRED &-&40.6&34.7&28.4  &-&39.4&36.5&31.1 \\
 ReCoSa &59.4&-&39.3&35.7    &60.6&-&42.7&35.0 \\ 
 BART &59.4&-&39.3&35.7    &60.6&-&42.7&35.0 \\ 
 GNN &65.3&60.7&-&38.5       &63.5&57.34&-&38.4 \\
 HGNN &\bf{71.6}&64.3&61.5&- &\bf{68.9}&{65.00}&{61.6}&- \\
\bottomrule
\end{tabular}}
\caption{Preference test (\%) between any two models in `Two-Sources' setting.
}\label{human_evaluation_Preference} 
\end{table}

\begin{table}[t]
\centering
\setlength{\tabcolsep}{0.9mm}{
\begin{tabular}{l|l|rccccccc}
\toprule
\#&Models & PPL $\downarrow$ & BLEU $\uparrow$ & Dist-1 $\uparrow$ &Dist-2 $\uparrow$ &Avg.$\uparrow$ &Ext.$\uparrow$ &Gre.$\uparrow$&W-avg. $\uparrow$\\\hline
1&HGNN &{96.34} &{3.08} &{5.08}  &{9.58}&{0.597}&{0.398}&{0.442}&{34.72}\\
2&- facial expressions &113.94 &2.17 &2.31  &5.95&0.568&0.367&0.413&31.54 \\
3&- audio              &114.94 &2.23 &2.27  &6.18&0.568&0.369&0.417&31.04 \\
4&- emotion            &111.42 &2.09 &2.18  &6.37&0.571&0.364&0.419&30.18\\
5&- all speakers       &100.95 &2.50 &2.08  &8.60&0.573&0.371&0.418&32.92\\
6&- current speaker &101.91 &2.62 &4.55 &8.87&0.587&0.388&0.431&33.64\\
\bottomrule
\end{tabular}}
\caption{Ablation study on MELD. `- current speaker' means removing it from the decoder while maintaining it in HGNN.
}\label{ablation} 
\end{table}

\begin{table}[t]
\centering
\setlength{\tabcolsep}{0.9mm}{
\begin{tabular}{l|l|rccccccc}
\toprule
\#&Models & PPL $\downarrow$ & BLEU $\uparrow$ & Dist-1 $\uparrow$ &Dist-2 $\uparrow$ &Avg.$\uparrow$ &Ext.$\uparrow$ &Gre.$\uparrow$&W-avg. $\uparrow$\\\hline
1&Vanilla HGNN &{118.89} &{1.89} &{1.79}  &{5.15}&{0.552}&{0.346}&{0.409}&{29.47}\\
2& 1 + facial expressions &108.17 &2.32 &3.31  &6.95&0.568&0.367&0.413&31.54 \\
3& 2 + audio              &107.61 &2.43 &3.89  &7.18&0.572&0.371&0.417&31.99 \\
4& 3 + emotion            &103.77 &2.49 &3.98  &7.37&0.583&0.379&0.425&32.92\\
5& 4 + all speakers       &101.91 &2.62 &4.55 &8.87&0.587&0.388&0.431&33.64\\
6& 5 + current speaker    &{96.34} &{3.08} &{5.08}  &{9.58}&{0.597}&{0.398}&{0.442}&{34.72}\\
\bottomrule
\end{tabular}}
\caption{Ablation study of adding one type of knowledge at a time on MELD. For example, row 2 (`1 + facial expressions') means adding facial expressions under row 1 setting (Vanilla HGNN).
}\label{ablation_all} 
\end{table}

\begin{table}[t]
\centering
\setlength{\tabcolsep}{0.9mm}{
\begin{tabular}{l|l|c|c}
\toprule
Data Sources&\multirow{ 1}{*}{Methods} & \multicolumn{1}{c|}{MELD} &  \multicolumn{1}{c}{DailyDialog} \\\cline{1-4}

\multirow{6}{*}{\tabincell{c}{Two Sources}}
&Emo-HRED  &31.03  &79.67 \\
&ReCoSa   &32.19  &81.60 \\
&BART &35.25 &82.96 \\
&GNN (Ours)  &32.35  &81.84 \\
&HGNN (Ours) &{33.52} &{82.70} \\
&HGNN+BART (Ours) &{36.73} &\bf{83.88} \\\hline
\multirow{5}{*}{\tabincell{c}{Four Sources}}
&ReCoSa &33.41 &- \\
&BART &36.71 &- \\
&GNN (Ours) &34.49 &- \\
&HGNN (Ours)  &{36.28} &- \\
&HGNN+BART (Ours)  &{37.58} &- \\\hline
\multirow{5}{*}{\tabincell{c}{Five Sources}}
&ReCoSa &33.72 &- \\
&BART &37.07 &- \\
&GNN (Ours) &34.83 &- \\
&HGNN (Ours)  &{36.55} &- \\
&HGNN+BART (Ours)  &\bf{38.12} &- \\
\bottomrule
\end{tabular}}
\caption{The `W-avg.' F1-score (\%) for the \emph{Emotion Predictor} of different methods on test sets. 
}\label{emotion_classification_wavg} 
\end{table}

\begin{table}[t]
\centering
\setlength{\tabcolsep}{0.9mm}{
\begin{tabular}{l|l|rccccccc}
\toprule
Datasets&Models & PPL $\downarrow$ & BLEU $\uparrow$ & Dist-1 $\uparrow$ &Dist-2 $\uparrow$ &Avg.$\uparrow$ &Ext.$\uparrow$ &Gre.$\uparrow$&W-avg. $\uparrow$\\\hline
\multirow{2}{*}{\tabincell{c}{MELD}}
&HGNN &{96.90} &{2.80} &{4.84}  &{9.15}&{0.593}&{0.394}&{0.437}&{34.49}\\
& w/ golden emotion &95.49 &2.84 &5.13  &9.69 &0.598&0.398&0.445&- \\\hline
\multirow{2}{*}{\tabincell{c}{DailyDialog}}
&HGNN   &{76.59} &{0.75} &{0.22} &{0.47} &{0.641}&{0.418}&{0.510}&{71.16} \\
&w/ golden emotion  &76.14 &0.91 &0.35  &0.64&0.661&0.423&0.512&-\\
\bottomrule
\end{tabular}}
\caption{Model results with given golden emotion labels on MELD and DailyDialog.
}\label{upper_bound_res} 
\end{table}
\begin{table}[t]
\centering
\setlength{\tabcolsep}{0.8mm}{
\begin{tabular}{l|l|rccccccc}
\toprule
Datasets&Models & PPL $\downarrow$ & BLEU $\uparrow$ & Dist-1 $\uparrow$ &Dist-2 $\uparrow$ &Avg.$\uparrow$ &Ext.$\uparrow$ &Gre.$\uparrow$&W-avg. $\uparrow$\\\hline
\multirow{4}{*}{\tabincell{c}{MELD}}
&HGNN &{96.90} &{2.80} &{4.84}  &{9.15}&{0.593}&{0.394}&{0.437}&{34.49}\\
& w/o gate in Eq.(\ref{gate1}) &98.48 &2.64 &4.76  &8.71 &0.591&0.390&0.428&34.25 \\
& w/ ECM &89.31 &2.86 &4.99  &9.98&0.595&0.398&0.433&34.54 \\
&w/ EmoDS  &85.84 &2.94 &5.32  &10.78&0.599&0.405&0.445&34.88\\\hline
\multirow{4}{*}{\tabincell{c}{DailyDialog}}
&HGNN   &{76.59} &{0.75} &{0.22} &{0.47} &{0.641}&{0.418}&{0.510}&{71.16} \\
& w/o gate in Eq.(\ref{gate1}) &77.82 &0.70 &0.21  &0.41&0.635&0.409&0.503&71.04 \\
& w/ ECM &73.19 &1.21 &1.13  &1.97&0.648&0.426&0.517&71.45 \\
&w/ EmoDS  &71.42 &1.30 &1.58  &2.23&0.671&0.433&0.519&71.38\\

\bottomrule
\end{tabular}}
\caption{Results of our model equipped with another two emotional decoders (\emph{i.e.}, ECM~\citep{zhou2018emotional} and EmoDS~\citep{song-etal-2019-generating}).
}\label{upper_bound_res_decoder} 
\end{table}

\subsection{Human Evaluation Results} \label{human_evaluation_results}

Results of content quality, emotion quality, and personality quality are shown in Table~\ref{human_evaluation_content}, ~\ref{human_evaluation_emotion} and ~\ref{human_evaluation_per}, respectively. The Fleiss’ kappa~\citep{fleiss1973equivalence} for measuring inter-rater agreement is included as well. All models have “moderate agreement” or “substantial agreement”. 

\paragraph{Content Quality.} As shown in Table~\ref{human_evaluation_content}, for the MELD dataset, our HGNN has more +2 ratings than comparison models on content level (t-test~\citep{koehn-2004-statistical}, {\em p}-value \textless \ 0.05) in different sources settings, showing that our model is better at generating coherent responses than others. For the DailyDialog dataset, our HGNN model also achieves significantly many +2 scores, suggesting the generalizability of our model, which is consistent with the automatic evaluation. 

\paragraph{Emotion Quality.}
As shown in Table~\ref{human_evaluation_emotion}, for the MELD dataset, our HGNN has more +2 ratings than comparison models on emotion level (t-test~\citep{koehn-2004-statistical}, {\em p}-value \textless \ 0.05) in different sources settings, showing that our model is better at generating emotional responses than others. Especially, in `Five Sources' setting, the HGNN model achieves further improvements on +2 metric, which is also far beyond other baselines. This demonstrates that our model can effectively infuse facial expressions, audios and speakers' personalities into the dialogue graph and generate accurate emotions. For the DailyDialog dataset, our HGNN model also obtains very obvious promotion on +2 metric, suggesting the generalizability of our model again, which is also consistent with the automatic evaluation. 

\paragraph{Personality Quality.}
As shown in Table~\ref{human_evaluation_per}, for the MELD dataset, our HGNN has more +2 ratings than comparison models on personality level (t-test~\citep{koehn-2004-statistical}, {\em p}-value \textless \ 0.05) in different sources settings, showing that our model is better at generating personality-preserving responses than others. In `Five Sources' setting, the HGNN model achieves further improvements on +2 metric, which is also far beyond other baselines. This demonstrates that our model can effectively infuse speakers' personalities into the dialogue graph and preserve consistent personalities. Due to no speaker information in the DailyDialog dataset, so we omit the personality quality evaluation. 

\paragraph{Preference Test.} 
Furthermore, we conduct preference test in Table~\ref{human_evaluation_Preference}. We can see that HGNN is signiﬁcantly preferred against other models (t-test, p-value \textless \ 0.05) in `Two Sources' scenario. Obviously, the coherent, emotional, and personality-preserving responses generated by our HGNN are more attractive to users than the responses produced by baselines.

\section{Analysis}
In this section, we conduct thorough and in-depth analysis about our multi-source knowledge and each components of our model. Firstly, to further investigate the effect of each knowledge of our multi-source knowledge, we conduct in-depth analysis in Table~\ref{ablation}. Secondly, to investigate the performance of our model on golden emotion, we replace the predicted emotion predicted by the \emph{Emotion Predictor} with golden emotion, and the results are shown in Table~\ref{upper_bound_res}. Thirdly, to probe the performance of the \emph{Emotion Predictor} of the encoder, we also conduct comparison results with some existing models, as shown in Table~\ref{emotion_classification_wavg}. Finally, to further enhance our model performance, we integrate two decoders into our model, i.e., ECM~\citep{zhou2018emotional} and EmoDS~\citep{song-etal-2019-generating}, which are originally designed for emotion-controlled systems. Furthermore, we present two examples to 
intuitively show how our model works well, as shown in Figure~\ref{fig:case_analysis}.

\subsection{Ablation Study for Multi-source Knowledge}
Table~\ref{ablation} and Table~\ref{ablation_all} shows the results of the ablation study about heterogeneous nodes in HGNN. The experiments in Table~\ref{ablation} are removing one at a time. The experiments in Table~\ref{ablation} are adding one at a time. 

Row 1 of Table~\ref{ablation} or row 6 of Table~\ref{ablation_all} is the results of our full model, which incorporates not only the dialog history but also its emotion flow, the facial expressions (\emph{i.e.}, image), the audio, and the speakers. Rows 2$\sim$4 of Table~\ref{ablation} are the results of removing the information of corresponding nodes. From the results, we can conclude that after removing each heterogeneous node information, the response quality of content and emotion is differently lowered, suggesting that various information of heterogeneous nodes is crucial for understanding content, and emotional perception and expression (row 1 vs. rows 2$\sim$6), especially for emotion flow (row 3). 

From Table~\ref{ablation_all}, we find that each source knowledge makes substantial contributions to performance, \emph{i.e.}, adding one type of knowledge at a time can improve final performance, proving the importance of each source knowledge. This also sheds lights on how performance can be efficiently improved by infusing multi-source knowledge.

\subsection{Effect of the Model with Golden Emotion Labels} 
To test the performance of our model with golden emotion labels, we conduct such experiments in Table~\ref{upper_bound_res}. We can see that our model achieves better results than that with predicted emotion labels, showing the generalizability of our model on infusing either predicted emotion labels or golden emotion labels. This also demonstrates that the golden emotion labels can indeed further enhance the emotion expression and thus improve the model performance.

\subsection{\emph{Emotion Predictor} of the Encoder}
We also evaluate the performance of the \emph{Emotion Predictor} from the \emph{Heterogeneous Graph-Based Encoder} to show whether the encoder can fully perceive emotions from multi-source knowledge and predict suitable emotions. The results shown in Table~\ref{emotion_classification_wavg} suggest that our encoder can more effectively perceive emotions from multi-source knowledge than baselines, and therefore predict more accurate emotions for the decoder.


\subsection{Integrating Other Two Emotional Decoders} 
To further enhance the performance of our model, we integrate such two emotional decoders into our framework, i.e., ECM~\citep{zhou2018emotional} and EmoDS~\citep{song-etal-2019-generating}, which are designed to express user-input emotions. The ECM decoder exploits emotion embedding, internal memory and external memory to generate emotional responses. Rather than using the golden label for emotion, we use the embedding of the emotion label predicted by our \emph{Emotion Predictor}. The EmoDS decoder uses both explicit and implicit emotional representation to express the desired emotion in responses, where the desired emotion is predicted by our \emph{Emotion Predictor} rather than given golden emotion label in the original work.  

The results are shown in Table~\ref{upper_bound_res_decoder}. We can see that our model can obtain higher results after equipping with such decoders, \emph{i.e.}, ECM~\citep{zhou2018emotional} and EmoDS~\citep{song-etal-2019-generating}. And we also can find that our model with EmoDS can surpass that with ECM, suggesting that the powerful decoder EmoDS still works well based on our general framework. This also show the generalizability of our model.

\begin{figure}[t]
    \centering
    \includegraphics[width=0.75\textwidth]{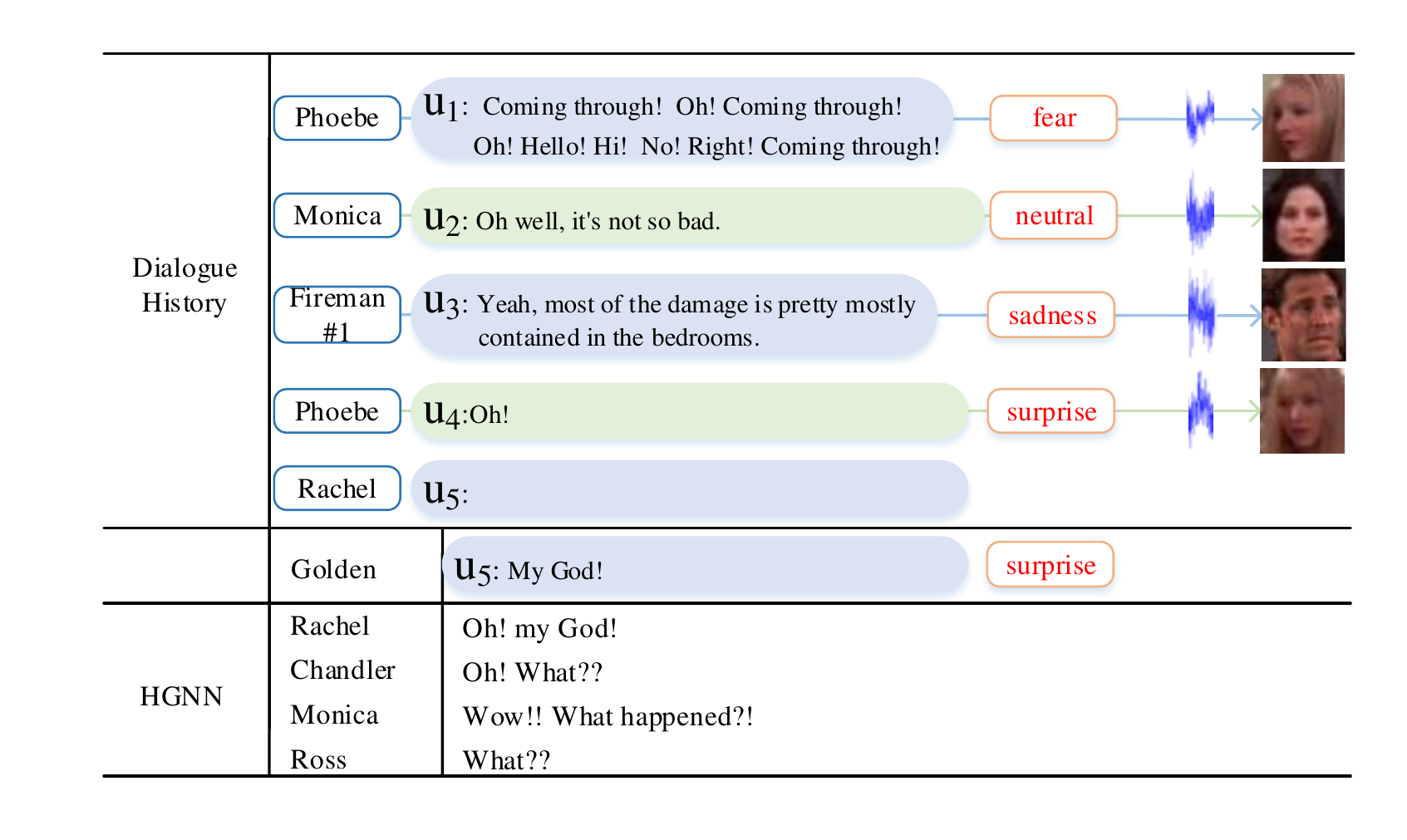}
    \caption{A dialogue example with different speaker inputs for measuring the speaker's personalities.
    }
    \label{fig:per}
\end{figure}

\begin{figure*}[h]
\centering
  \includegraphics[width = 1.0 \textwidth]{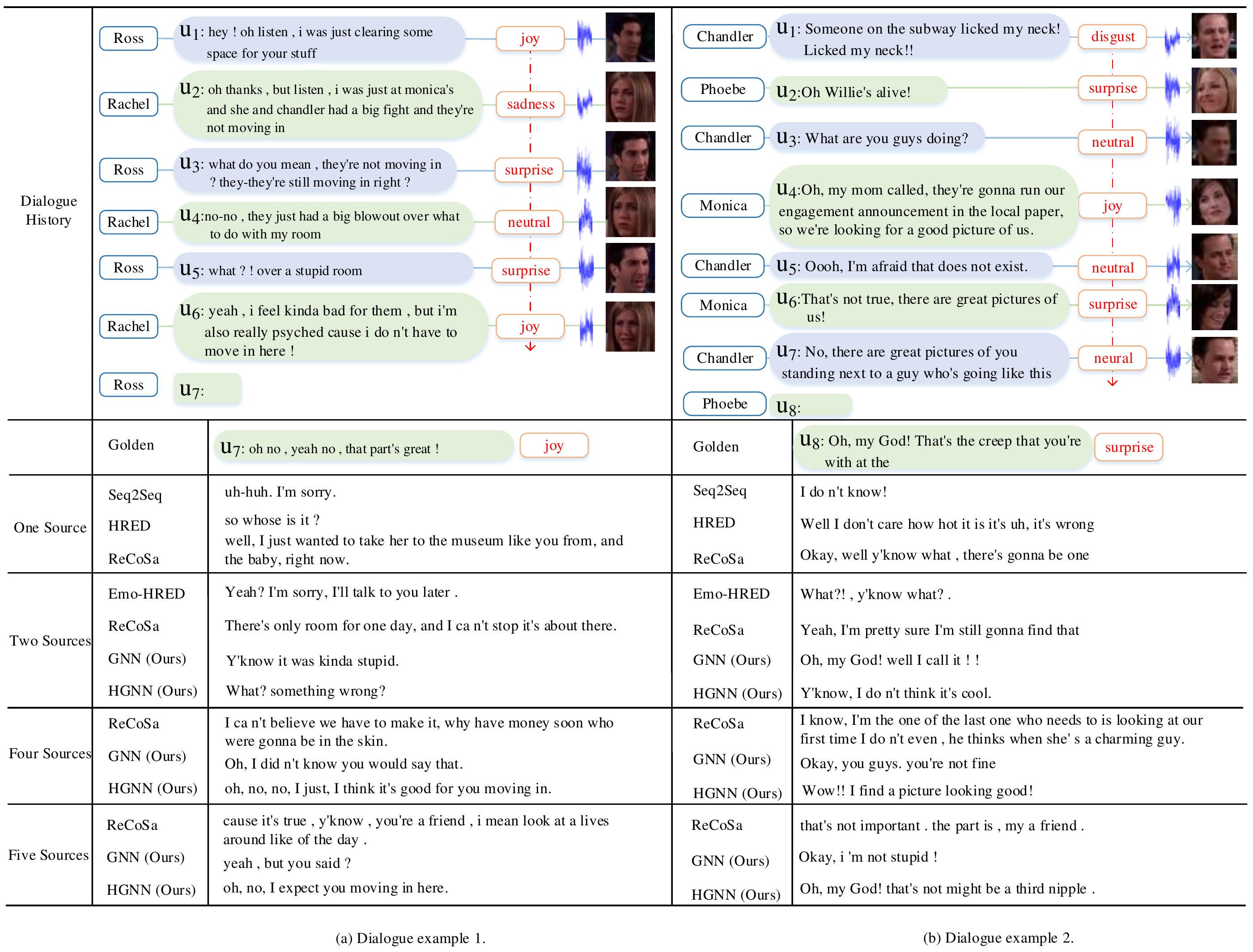}
\caption{Responses generated by all models without pretrained language model.} 
\label{fig:case_analysis}
\end{figure*}

\subsection{Case Analysis} 
\label{C_S}
To give an insight on whether the emotion of the generated response is expressed appropriately, we provide some examples from MELD test dataset in Figure~\ref{fig:case_analysis}. It shows that HGNN can generate coherent responses with suitable emotions.

Specifically, we can see that the methods in `One Source' setting generate some neutral and safe responses (\emph{e.g.}, ``uh-huh. I'm sorry.'' in example 1 or ``I don't know!'' in example 2 by Seq2Seq) due to lacking of additional emotion perception source. Although these methods in `Two Sources' can generate emotional responses (\emph{e.g.}, ``Yeah? I'm sorry, I'll talk to you later.'' by Emo-HRED in example 1 or ``Yeah, I'm pretty sure I'm still gonna find that'' by ReCoSa in example 2, may contain {\em anger} and {\em neutral} emotion, respectively.), the emotions of the responses are not suitable with the dialogue history. In `Four Sources'/`Five Sources' settings, although the ReCoSa and GNN generate more relevant responses after incorporating the facial expression and speakers' personalities, they can not express an appropriate emotion with the dialogue history. And the HGNN generates a coherent as well as emotional response, suggesting that our heterogeneous graph-based model can sufficiently understand the content and perception of emotion from various types of knowledge, and thus can effectively enhance the quality of responses.


Besides, we also provide one example to show speaker's personalities in Figure~\ref{fig:per}. In the same setting, we can observe that the generated responses given different speakers are usually personalized (\emph{i.e.} speaker’s personality) with the same or close meaning.

\section{Related Work}
Our work mainly involves dialog systems and heterogeneous graph neural network, so we divide the related work into two parts: dialog systems and heterogeneous graph neural network in natural language processing.
\subsection{Dialog Systems}
In the past few years, previous studies mainly focus on improving the content quality of dialogue~\citep{li-etal-2016-diversity,zhang-etal-2019-recosa,shuster-etal-2020-dialogue}, and only a little work pays attention to improving the emotion quality of dialogue~\citep{DBLP:journals/corr/abs-1812-08989,rashkin-etal-2019-towards}. Researchers firstly set user-input emotions directly for response generation (1), such as emotion-controllable conversation systems, which contain some representative work~\citep{zhou2018emotional,colombo-etal-2019-affect,zhou-wang-2018-mojitalk,huang-etal-2018-automatic,sun2018emotional}. 
To automatically learn emotions (2), researchers track the emotion flow of dialog history~\citep{Lubis2018ElicitingPE,lin-etal-2019-moel,li2019empgan,li2020empathetic,majumder-etal-2020-mime} and generate emotion-rich responses~\citep{zhong2019affect,asghar2018affective}, where they incorporate Valence, Arousal, and Dominance embeddings~\citep{warriner2013norms,mohammad-2018-obtaining} into their models to provide additional affective knowledge.
Besides, (3) multi-modal studies also have attracted much attentions in conversation systems such as video-grounded dialogue system~\citep{DBLP:journals/corr/abs-1901-03461} and visual question-answering~\citep{antol2015vqa}, aiming to answer human queries grounded a video or image. Finally, (4) as a close work~\citep{8578841}, it also considers facial expressions and conversation history to generate both text and face.

The differences between our model and those in (1) and (2) are: they only perceive emotions from text and ignore other source knowledge, \emph{e.g.}, the facial expressions and speakers' personalities, where the speaker information has been shown helpful for conversation systems~\citep{li-etal-2016-persona}. The deep difference from (1) is: we focus on automatically learning emotions and then expressing it rather than emotion-controllable conversation systems (\emph{i.e.} a limited scenario for expressing a user-input emotion).  Furthermore, we expect to construct a practical emotional conversation system, \emph{i.e.} infusing multi-source knowledge with heterogeneous graph neural network for improving emotional conversation, so we employ a simple and general decoder. Obviously, a powerful emotion-aware decoder can be integrated into our framework, such as~\citep{zhou2018emotional,colombo-etal-2019-affect,song-etal-2019-generating,shen-feng-2020-cdl}, which has been verified to further improve the emotional performance of our model. 
In contrast to tasks in (3), we mainly utilize multi-modal information for emotion perception and expression in response generation. Compared with (4), we mainly focus on generating emotional response by infusing multi-source knowledge rather than generating general text and face.

\subsection{Heterogeneous Graph for NLP}
Heterogeneous graph neural networks~\citep{7536145,10.1145/1516360.1516426} can deal with various types of nodes and edges and have more advantages than homogeneous graph neural networks~\citep{10.1145/3308558.3313562,zhang2020deep}. Its superiority has been verified in many natural language processing (NLP) tasks, such as graph representation learning~\citep{hong2019attentionbased}, reading comprehension~\citep{tu2019multi}, text classification~\citep{linmei2019heterogeneous,liang2020dependency,liang-etal-2021-iterative-multi} and extractive document summarization~\citep{wang2020heterogeneous}. Inspired by the success of heterogeneous graph neural network, we first introduce it to emotional conversation generation to gain a better understanding of content and fully perceive emotions from multi-source knowledge, and then produce a satisfactory response.

\section{Conclusion}

In this work, we firstly present a heterogeneous graph-based framework to understand dialogue content and fully perceive complex and subtle emotions from multi-source knowledge (i.e., the emotion flow hidden in dialogue history, facial expressions, audio, and personalities of speakers), and then predict suitable emotions for feedback. Finally, we devise an \emph{Emotion-Personality-Aware Decoder} to express it and thus generate a coherent, emotional, and personality-preserving response. We verify our model performance on two benchmark datasets, i.e., MELD and DailyDialog. Especially, we conduct extensive experiments on MELD to investigate the effect of multi-source knowledge on emotional dialogue.
Both automatic evaluation and human evaluation results demonstrate the effectiveness and generalizability of our model, which can be also easily adapted to different number of knowledge sources. And the further experimental analysis and case study also suggest our model can generate more attractive and satisfactory response than baselines. Besides, we also integrate other decoders, like ECM~\citep{zhou2018emotional} and EmoDS~\citep{song-etal-2019-generating}, into our framework to further enhance the model performance. Finally, based on the up-to-date text generator BART, our model still achieves consistent improvement. This shows the generalizability of our model again.


\section*{Acknowledgments}
Liang, Zhang, Chen and Xu are supported by the National Key R\&D Program of China (2019YFB1405200) and the National Nature Science Foundation of China (No. 61976015, 61976016, 61876198 and 61370130). Liang is supported by 2021 Tencent Rhino-Bird Research Elite Training Program. We thank all anonymous reviewers for their valuable suggestions.

\bibliographystyle{plainnat}
\bibliography{AIJ}

\end{document}